\documentclass{article}

\usepackage{times}
\usepackage{graphicx} 
\usepackage{subfigure}
\usepackage{amssymb,amsfonts,amsmath}
\usepackage{float}

\usepackage{natbib}

\usepackage{algorithm}
\usepackage{algorithmic}

\usepackage{hyperref}

\usepackage[accepted]{icml2016mine}

\icmltitlerunning{Efficient Representation of Low-Dimensional Manifolds using Deep Networks}

\newcommand{\av}{{\mathbf a}}
\newcommand{\ao}{{\mathbf a_0}}
\newcommand{\bo}{{\mathbf b_0}}
\newcommand{\bv}{{\mathbf b}}
\newcommand{\h}{{\mathbf h}}
\newcommand{\p}{{\mathbf p}}
\newcommand{\po}{{\mathbf p_0}}
\newcommand{\uu}{{\mathbf u}}
\newcommand{\vv}{{\mathbf v}}
\newcommand{\w}{{\mathbf w}}
\newcommand{\x}{{\mathbf x}}
\newcommand{\y}{{\mathbf y}}
\newcommand{\C}{{\cal C}}
\newcommand{\R}{\mathbb R}

\newcommand{\longversion}[1]{}

\begin{document}

\twocolumn[
\icmltitle{Efficient Representation of Low-Dimensional Manifolds using Deep Networks}

\icmlauthor{Ronen Basri}{ronen.basri@weizmann.ac.il}
\icmladdress{Weizmann Institute of Science, Rehovot, 76100 ISRAEL}
\icmlauthor{David Jacobs}{djacobs@cs.umd.edu}
\icmladdress{Department of Computer Science and UMIACS, University of Maryland, College Park, MD 20742 USA}

\icmlkeywords{Deep network, low-dimensional manifolds}

\vskip 0.3in
]

\begin{abstract}
We consider the ability of deep neural networks to represent data that lies near a low-dimensional manifold in a high-dimensional space.  We show that deep networks can efficiently extract the intrinsic, low-dimensional coordinates of such data.  We first show that the first two layers of a deep network can exactly embed points lying on a {\em monotonic chain}, a special type of piecewise linear manifold, mapping them to a low-dimensional Euclidean space.  Remarkably, the network can do this using an almost optimal number of parameters. We also show that this network projects nearby points onto the manifold and then embeds them with little error.  We then extend these results to more general manifolds.

\end{abstract}

\section{Introduction}


Deep neural networks have achieved state-of-the-art results in a variety of tasks. This remarkable success is not fully explained, but one possibility is that their hierarchical, layered structure may allow them to capture the geometric regularities of commonplace data. We support this hypothesis by exploring ways that networks can handle input data that lie on or near a low-dimenisonal manifold. In many problems, for example face recognition, data lie on or near manifolds that are of much lower dimension than the input space~\cite{turk1991eigenfaces,basri2003lambertian,lee2003video}, and that represent the intrinsic degrees of variation in the data. 

\begin{figure}[t]
\begin{center}
\includegraphics[width=3.3cm]{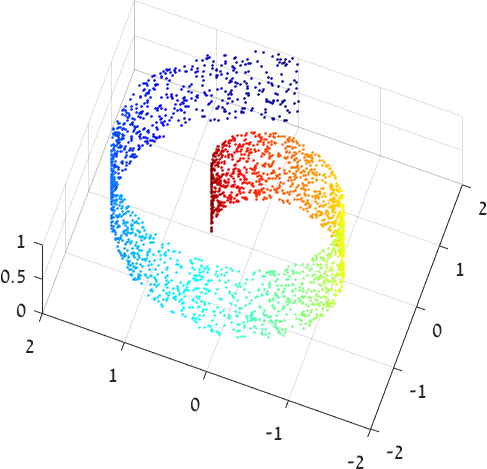}
\includegraphics[width=5.8cm]{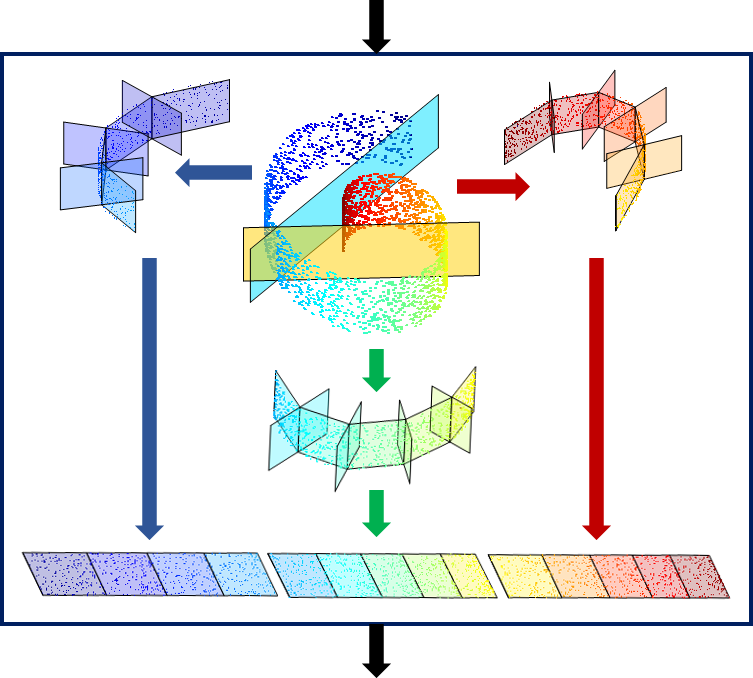}
\includegraphics[width=4.cm]{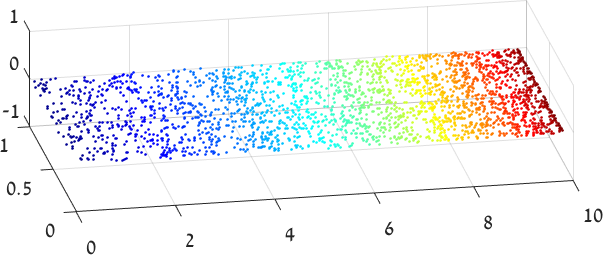}
\end{center}
\caption{We illustrate the embedding of a manifold by a deep network using the famous Swiss Roll example (top).  Dots represent color coded input data.   In the center, the data is divided into three parts using hidden units represented by the yellow and cyan planes.  Each part is then approximated by a monotonic chain of linear segments.  Additional hidden units, also depicted as planes, control the orientation of the next segments in the chain.  A second layer of the network then flattens each chain into a 2D Euclidean plane, and assembles these into a common 2D representation (bottom).}
\label{fig:Overview}
\end{figure}

We study the representational power of deep networks when applied to manifold data.  We demonstrate that the initial layers of networks can take inputs that lie in a manifold in a high-dimensional space, approximate this manifold with piecewise linear functions, and economically output the coordinates of these points embedded in a low-dimensional Euclidean space. In fact, each new linear segment approximating the manifold can be represented by a single additional hidden unit, leading to a representation of manifold data that in some cases is nearly optimal in the number of parameters of the system.  This means that subsequent layers of a deep network could build upon these early layers, operating in lower dimensional spaces that more naturally represent the input data.
It is beyond the scope of this paper to study the problem of training networks to build these representations.  However, our results describe novel representations that might be sought in existing networks, or that might suggest new architectures for networks.  Moreover, we feel that these results provide intuitions about the role that individual units of a network can play in shaping the function that it is computing.

We first show how this embedding can be done efficiently for manifolds consisting of {\em monotonic chains} of linear segments.  We then show how these primitives can be combined to form linear approximations for more complex manifolds. This process is illustrated in Figure \ref{fig:Overview}.  We further show that when the data lies sufficiently close to their linear approximation, the error in the embedding will be small.  Our constructions will use a feed-forward network with rectified linear unit (RELU) activation. We consider fully connected layers, although the treatment of complex manifolds that are divided into pieces (e.g., of monotonic chains) will be modular, resulting in many zero weights.

\section{Prior Work}

Realistic learning problems, e.g., in vision applications and speech processing, involve high dimensional data. Such data is often governed by many fewer variables, producing manifold-like sub-structures in a high dimensional ambient space. A large number of dimensionality reduction techniques, such as principle component analysis (PCA)~\cite{PCA}, multi-dimensional scaling~\cite{MDS}, Isomap~\cite{Isomap}, and local linear embedding (LLE)~\cite{LLE}, have been introduced.  An underlying {\em manifold assumption}, which states that different classes lie in separate manifolds, has also guided the design of clustering and semi-supervised learning algorithms~\cite{DiffusionMap,LaplacianEigenmaps,Weston,Mobahi}.

A number of recent papers examine properties of neural nets in light of this manifold assumption. Specifically, Rifai et al.~(\citeyear{Rifai}) trained a contractive auto-encoder to represent an atlas of manifold charts. Shaham et al.~(\citeyear{Shaham}) demonstrate that a 4-layer network can efficiently represent any function on a manifold through a trapezoidal wavelet decomposition. In both, each chart is represented independently, requiring the representation of an independent projection to map the input space onto each chart. We show that for monotonic chains we can reduce the size of the representation to near optimal by exploiting geometric relations between neighboring projection matrices, so that an additional chart requires only a single hidden unit.   

Another family of networks attempt to learn a ``semantic'' distance metric for training pairs, often by using a siamese network~\cite{Salakhutdinov,Chopra,Hadsell,YiLei,HuangLang}. These assume that the input space can be mapped non-linearly by a network to produce the desired distances in a lower dimensional  feature space.   \cite{Sapiro} shows that even a feed-forward neural network with random Gaussian weights embeds the input data in an output space while preserving distances between input items. They further suggest that training may  improve the embedding quality.


Another outstanding question is to what extent deep networks can  represent data or handle classification problems more efficiently than shallow networks with a single hidden layer. Earlier work showed that shallow networks are universal approximators~\cite{cybenko1989approximation}. However, recent work demonstrates that deep networks can be exponentially more efficient in representing certain functions~\cite{Bianchini,Telgarsky,Eldan,Delalleau,Montufar,Cohen}.  On the other hand, \cite{ba2014deep} show empirically that in many practical cases a shallow network can be trained to mimic the behavior of a deep network.  Our construction does not produce exponential gains, but does show that the early layers of a network can efficiently reduce the dimensionality of data that feeds into later layers.






\label{sec:prior_work}

\section{Monotonic Chains of Affine Subspaces}  \label{sec:monotonic_chains}

Our aim in this paper is to construct networks that can perform dimensionality reduction for data that lies on or near a manifold. We focus on feed-forward networks with RELU activation, i.e., $\max(x,0)$. Clearly the output of such networks are continuous, non-negative piecewise linear functions of their input. It is therefore natural to ask whether they can embed piecewise-linear manifolds in a low-dimensional Euclidean space both accurately and efficiently. In this section we construct such efficient networks for a class of manifolds that we call {\em monotonic chains of affine subspaces}, which are defined shortly. These will serve as building blocks for handling more general chains, as well as other sets of data, which can be decomposed into monotonic chains. Handling these more complex cases will require deeper networks. In subsequent sections we discuss these more complex manifolds and show in addition that our networks can be used to approximate data that is on or near non-linear manifolds.

\begin{figure}
\begin{center}
\includegraphics[width=3.7cm]{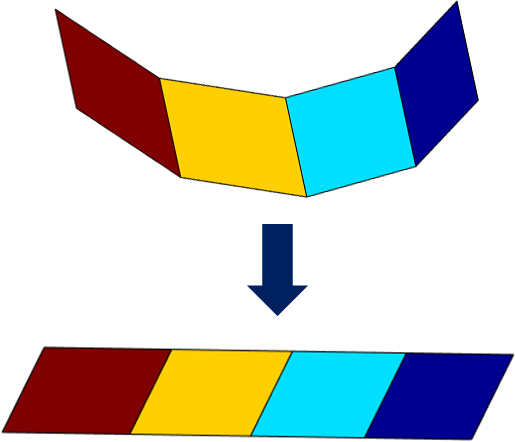}
\end{center}
\caption{A continuous chain of linear segments (above) that can be flattened to lie in a single low-dimensional linear subspace (bottom).}
\label{fig:Chain}
\end{figure}

We will consider the case of data lying in a chain of linear segments, denoted $\C=S_1 \cup ... \cup S_K$. Each segment $S_k$ ($1 \le k \le K$) in the chain is a portion of some $m$-dimensional affine subspace of $\R^d$, and the segments are connected to form a chain (Figure \ref{fig:Chain}). We suppose that every two subsequent segments $S_{k-1}$ and $S_k$ intersect, and that the intersection lies in an $(m-1)$-dimensional affine subspace. We further assume that these chains can be flattened so that they may be represented in $\R^m$. Note that any curve on $\C$ will be mapped to a curve of the same length in $\R^m$ on the flattened chain. 

We will next consider a special case of these chains which we call {\em monotonic}, and show that these can be handled using networks with two hidden layers. 

\begin{figure}
\begin{center}
\includegraphics[width=3.4cm]{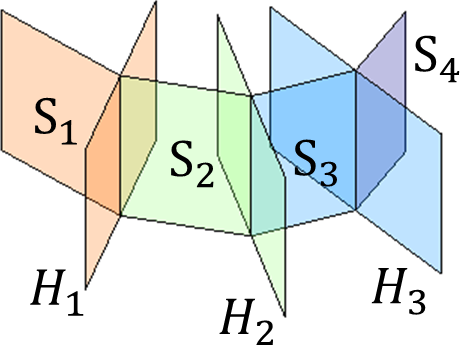}
\end{center}
\caption{A monotonic chain.  $S_k$ denotes the $k$'th segment in the chain. $H_k$ is a hyperplane that separates $S_1, ..., S_k$ from $S_{k+1}, ..., S_K$.}
\label{fig:MonotonicChain}
\end{figure}

{\bf Definition:} We say that a chain of $K$ affine subspaces is {\em monotonic} (see Figure \ref{fig:MonotonicChain}) when there exist a set of $K-1$ half-spaces, $H_1, H_2, ..., H_{K-1}$ such that $H_k$ is bounded by a hyperplane that contains the intersection of $S_k$ and $S_{k+1}$, and $S_{k+1}, S_{k+2}, ..., S_K \subset H_k$ while $S_1, S_{2}, ..., S_k \subset H^C_k$, where $H^C_k$ is the complement of $H_k$.  Intuitively, each of the half-spaces divides the chain into two connected pieces at the boundary of each linear segment.  We can consider each half-space to represent a hidden unit that is {\em active} (i.e., non-zero) over a subset of the regions.  With a monotonic chain, the set of active units grows monotonically, so that, $H_{k+1} \subseteq H_k$.  Additionally, we can always define some units that are active over all the regions.

Below we show that monotonic chains can be embedded efficiently by networks with two layers of weights. These networks have $d$ units in the input layer, a hidden layer with $\kappa$ units that encodes the structure of the manifold (with $\kappa=K+m-1$ is a function of the manifold complexity), and an output layer with $m$ units. Denote the weights in the first layer by a $\kappa \times d$ matrix $A$ and further use a bias vector $\ao \in \R^\kappa$. The second layer of weights is captured by a $\kappa \times m$ matrix $B$. The total number of weights in these two layers is $(d+m+1)(K+m-1)$. This two layer network maps a point $\x \in \R^d$ to the embedding space $\R^m$ through
\[ \uu = B[A\x+\ao]_+ \]
where $[.]_+$ denotes the RELU operation. For now we do not use a bias or RELU in the second level, but those will be used later when we discuss more complex manifolds.

A simple example of a manifold that can be represented efficiently with a neural network occurs when the data lies in a single $m$-dimensional affine subspace of $\R^d$. Embedding can be done in this case with just one layer, with the matrix $A$ of size $m \times d$ containing in its rows a basis parallel to the affine space. RELU is not needed, but if required we can set the bias $\ao$ accordingly to map all the feasible data points to non-negative coordinates.

A simple way to extend this example to handle chains is by encoding each linear segment separately. Such encoding will require $mK$ units in addition to units that use RELU to separate each segment from the rest of the segments. A related representation was used, e.g., in \cite{Shaham}. Below we show that monotonic chains can be encoded much more efficiently.

We next show how to construct the network (i.e., set the weights in $A$, $\ao$, and $B$) to encode monotonic chains. Below we use the notation $A^{(k)}$ to denote the matrix formed by the first $k$ \textit{rows} of $A$, $\ao^{(k)}$ is the vector containing the first $k$ entries of $\ao$, and $B^{(k)}$ the matrix including the first $k$ \textit{columns} of $B$. Therefore $B^{(k)}[A^{(k)}\x+\ao^{(k)}]_+$ will express the output of the network when only the first $k$ hidden units are used. These will be set to recover the intrinsic coordinates of points in the first $k$ segments in $\C$; RELU ensures that subsequent hidden units do not affect the output for points in these segments.

For the construction we consider the pull-back of the standard basis of $\R^m$ on the chain, producing a geodesic basis to the manifold that is expressed by a collection of $d \times m$ column-orthogonal matrices $X^{(1)}, X^{(2)}, ..., X^{(K)}$.   Each matrix provides an orthogonal basis for one of the segments.

We will construct the network inductively. Suppose $k=1$. We set $A^{(1)}=X^{(1)\,T}$, $B^{(1)}=I$, and set $\ao^{(1)}$ so that for all $\x \in {\cal C}$ all the components of $A^{(1)} \x + \ao^{(1)}$ are non-negative. Clearly, $B^{(1)}A^{(1)}=X^{(1)\,T}$ is an orthogonal projection matrix and $B^{(1)}A^{(1)}X^{(1)}=I$.  This shows that the network projects the orthonormal basis for the first segment into $I$, an orthonormal basis in ${\cal R}^m$.  
Next we will show that $B^{(k)}A^{(k)}X^{(k)}=I$ for all $k$.  This implies that $B^{(k)}A^{(k)}x = X^{(k)\,T}x$, so there is no distortion in the projection. This will show that the network extends this basis throughout the monotonic chain in a consistent way.

Next, suppose we used $m+k-2$ units to construct $A^{(k-1)}$, $\ao^{(k-1)}$, and $B^{(k-1)}$ for the first $k-1 \ge 1$ segments. (For notational convenience we will next omit the superscript $k-1$ for these matrices and vectors, so $A=A^{(k-1)}$, etc.) We will now use those to construct $A^{(k)}$, $\ao^{(k)}$, and $B^{(k)}$.  We do so by adding a node to the first hidden layer. The weights on the incoming edges to this node will be encoded by appending a row vector $\av^T \in \R^d$ to $A$ and a scalar $a_0$ to $\ao$, and the weights on the outgoing edges will be encoded by appending a column vector $\bv \in \R^m$ to $B$. Our aim is to assign values to these vectors and scalar to extend the embedding to $S_{k}$.

By induction we assume that any $\tilde\x \in S_1 \cup ... \cup S_{k-1}$ is embedded with no distortion to $\R^m$ by
\[ \tilde\uu = B[ A{\tilde\x}+\ao ]_+, \]
and that $BAX=I$.
By monotonicity we further assume that $S_{k-1} \cap S_k$ is $m-1$ dimensional and there exists a hyperplane $H$ with normal $\h \in \R^d$ that contains this intersection with $C-(S_1\cup...\cup S_{k-1})$ lying completely on the side of $H$ in the direction of $\h$, while $S_1 \cup... \cup S_{k-1}$ lies on the opposite side of $H$. We then set $\av=\h$ and set $a_0$ so that $\av^T \bar\x + a_0=0$ for any point $\bar x \in S_{k-1} \cap S_k$. (This is well defined since $\h$ is orthogonal to $S_{k-1} \cap S_k$.)

To determine $\bv$, we first rotate the bases $X^{(k-1)}$ (referred to as $X$ below) and $X^{(k)}$ by a common, $m \times m$ matrix $R$, i.e., $Y=XR$ and $Y^{(k)}=X^{(k)}R$ so that $Y=[\w,\y_2,...,\y_m]$ and $Y^{(k)}=[\vv,\y_2,...,\y_m]$ with $\y_2,...,\y_m$ providing an orthogonal basis parallel to $S_{k-1} \cap S_k$. (This is equivalent to rotating the coordinate system in the embedded space and then pulling-back to the manifold.) Note that by the induction assumption $BAYR^T=I$. We next aim to set $\bv$ so that $B^{(k)}A^{(k)}X^{(k)}=I$. We note that
\[ B^{(k)}A^{(k)}X^{(k)} = B^{(k)}A^{(k)}Y^{(k)}R^T =
(BA+\bv\av^T)Y^{(k)}R^T. \]
We aim to set $\bv$ so that $(BA+\bv\av^T)Y^{(k)}R^T=I=BAYR^T$.
Consider this equality first for the common columns $\y_2,...,\y_m$ of $Y$ and $Y^{(k)}$. These columns are parallel to $S_{k-1} \cap S_k$, so that $\av^T \y_j=0$ for $2 \le j \le m$, implying equality for any choice of $\bv$. Consider next the left-most column of $Y$ and $Y^{(k)}$, denoted respectively $\w$ and $\vv$, we get
\[ (BA+\bv\av^T)\vv=BA\w. \]
This is satisfied if we set
\[ \bv=\frac{1}{\av^T\vv}BA(\w-\vv). \]

We have constructed $\bv$ so that the segments are embedded with consistent orientations.  We now show that they are also translated properly by $\ao$, to create a continuous embedding.   Consider a point $\x \in S_k$. Denote by $\bar\x$ its projection onto $S_{k-1} \cap S_k$, so that $\x=\bar\x+\beta \vv$ for a scalar $\beta$. Denoting the embedded
coordinates of $\x$ by $\uu$,
\[ \uu = B^{(k)}(A^{(k)}\x+\ao^{(k)} ). \]
We want to verify that as $\beta$ tends to 0 $\uu$ will coincide with the embedding of $\bar \x$ due to $S_{k-1}$, i.e.,
\[ \bar \uu=B(A\bar \x + \ao). \]
Due to the construction of $B^{(k)}$, $A^{(k)}$, and $\ao^{(k)}$
\[ \uu = (BA+\bv\av^T)\x+B\ao+a_0\bv. \]
Replacing $\x=\bar\x+\beta \vv$ we obtain
\[ \uu = (BA+\bv\av^T)\bar\x+\beta(BA+\bv\av^T)\vv+B\ao+a_0\bv. \]
Since $\av=\h$, $\av^T \bar\x+a_o=0$ and we get
\[ \uu = B(A\bar\x+\ao)+\beta(BA+\bv\av^T)\vv , \]
which coincides with $\bar\uu$ when $\beta \rightarrow 0$, implying that the embedding is extended continuously to $S_k$. Note that by construction $\av^T \y + a_0 \le 0$ for all $\mathbf{y} \in S_1 \cup ... \cup S_{k-1}$ so RELU ensures that the embedding of the these segments will not be affected by the additional unit.

Finally, we note that the proposed representation of monotonic chains with a neural network is very efficient and uses only few parameters beyond the degrees of freedom needed to define such chains. In particular, the definition of a chain requires specifying $m$ basis vectors in $\R^d$ for one linear segment (exploiting orthonormality these require $m(d-(m+1)/2)$ parameters), with each additional segment specified by a 1D direction for the new segment (a unit vector in $R^d$ specified by $d-m-1$ parameters) and a direction in the previous segment to be replaced (specified by a unit vector in $\R^m$, i.e. $m-1$ parameters). The total number of degrees of freedom of a chain is therefore $N=m(d-(m+1)/2)+(K-1)(d-2)$. This is the minimum possible number of parameters required to specify a monotonic chain.  Our construction requires $N'=(K+m+1)(d+m+1)$ parameters. Specifically, note that for any choice of parameters $K, d, m > 0$
\[ N \ge (K+m-1)(d-m-2). \]
We therefore obtain that
\[ \frac{N'}{N} \le \left(1 + \frac{2}{K+m-1}\right)\left(1+\frac{2m+3}{d-m-2}\right). \]
Assuming $d,K+m >> 1$ we get
\[ \frac{N'}{N} \lessapprox 1 + \frac{2m}{d-m}.\]
Since we normally expect that the dimension of the input space will be much greater than the dimension of the manifold, this ratio will be close to 1, which would be optimal.


\section{Error Analysis}

We now consider points that do not lie exactly on the monotonic chain.  We expect this to happen due to noise, or because we are approximating a non-linear manifold with piece-wise linear segments.  Let $\p_0$ be a point that is on the segment $S_j$, but that is then perturbed by some small noise vector, $\delta$, that is perpendicular to $S_j$, to produce the point $\p = \p_0 + \delta$.  Ideally, the network would represent $\p$ using the coordinates of $\p_0$.  In effect, the network would project all points onto the monotonic chain.  We now analyze the error that can occur in this projection.  Our analysis assumes that $\| \delta \|$ is small enough that $\p$ and $\p_0$ lie in the same region; that is, that they are both on the same side of all hyperplanes defined by the hidden units.

We first show in Section \ref{sec:worst} that for an arbitrary monotonic chain, this error can be unbounded.  While this sounds bad, we then show in Section \ref{sec:bounds} that this can only happen when the hyperplanes that separate the monotonic chain into segments must be poorly chosen, in some sense.  We show that in many reasonable cases the error is bounded by $\delta$ times a small constant.  

\subsection{Worst-case error}
\label{sec:worst}

\begin{figure}
\begin{center}
\includegraphics[width=6.5cm]{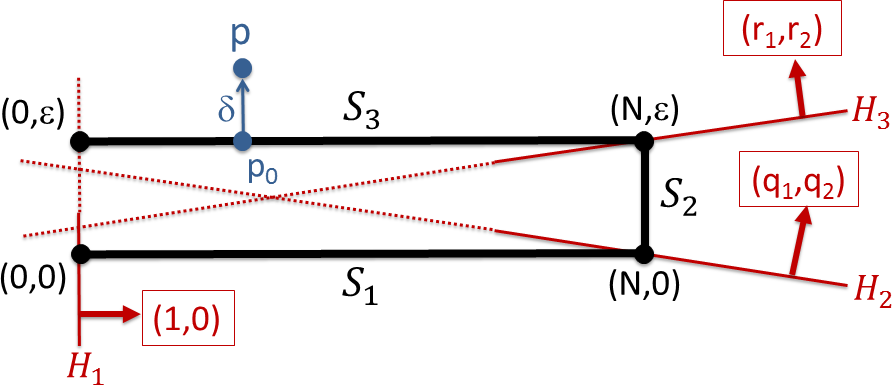}
\end{center}
\caption{In black, we show a 1D monotonic chain with three segments.  In red, we show three hidden units that flatten this chain into a line.  Note that each hidden unit corresponds to a hyperplane (in this case, a line) that separates the segments into two connected components.   The third hyperplane must be almost parallel to the third segment.  This leads to large errors for noisy points near $S_3$. }
\label{fig:WorstError}
\end{figure}

To show that the error can be unbounded, we consider a simple case in which the piecewise linear manifold consists of three connected 1D line segments, $S_1, S_2$ and $S_3$, with 2D vertices respectively of $(0,0)$ and $(N,0)$, $(N,0)$ and $(N,\epsilon)$, and $(N,\epsilon)$ and $(0,\epsilon)$.  $N$ is very large, and $\epsilon$ is very small (see Figure~\ref{fig:WorstError}).  Since three segments compose a 1D manifold, three hidden units defining three hyperplanes, $H_1, H_2$ and $H_3$ (lines) will be needed to represent the manifold.  In addition, a single output unit will sum the results of these units to produce the geodesic distance from the origin to any point on the three segments.

Using our construction in Section~\ref{sec:monotonic_chains} we get the embedding $f(\p) = B[ A{\p}+\ao ]_+$ with 
\begin{eqnarray*}
B&=&\left(1,\frac{1}{q_2},-\frac{1}{r_1}\left(2+\frac{q_1}{q_2}\right)\right),\\
A&=&\left(\begin{array}{cc}1&0\\q_1&q_2\\r_1&r_2\end{array}\right), ~~~
\ao=\left(\begin{array}{c}0\\ q_3\\ r_3\end{array}\right). 
\end{eqnarray*}
Note that the first row of $A$ uses the standard orthogonal projection $(x,y) \rightarrow x$; the two other rows of $A$ and $\ao$ separate the three segments with (1) $q_1,q_2>0$ and $q_1/q_2 \le \epsilon/N$ and $q_3=-q_1N$ set so that the separator $H_2$ goes through $(N,0)$, and (2) $r_1<0$, $r_2>0$ and $r_1/r_2 \ge -\epsilon/N$, and $r_3=-r_1N-r_2\epsilon$ set so that the separator $H_3$ goes through $(N,\epsilon)$. It can be easily verified that in this setup points on the first segment $(x,0)$, $0 \le x \le N$ are mapped to $x$, points $(N,y)$, $0 \le y \le \epsilon$ on the second segment are mapped to $N+y$, and points $(x,\epsilon)$, $0 \le x \le N$ on the third segment are mapped to $N+\epsilon+(N-x)$.

Ideally, we would want $\p$ to be embedded to the same point as $\po$. Let $E(\p)=f(\p)-f(\po)$. Clearly $E(\p)=B^{(k)}A^{(k)}\delta$. It can be readily verified that, under these conditions, when $\po \in S_1$ then $E(\p)=0$; when $\po \in S_2$ then $E(\p)=(1+q_1/q_2)\delta$, and when $\po \in S_3$ then $E(\p)=(1-(r_2/r_1)(2+q_2/q_1))\delta$. Therefore, there is no error in embedding $\p$ for $\po \in S_1$. The error in embedding $\p$ with $\po \in S_2$ is small and bounded (since $q_1/q_2 \le \epsilon/N$, assuming $\epsilon$ is small and $N$ is large), while the error in embedding $\p$ when $\po \in S_3$ can be huge since $-r_2/r_1 \ge N/\epsilon$.  In the next section we show that this can only happen when there is a large angle between a segment and the normal to the previous separating hyperplane.

\subsection{Bounds on Error}
\label{sec:bounds}


To show that this noise can often be quite limited, we will consider a class of monotonic chains in which the total curvature between all segments is less than or equal to some angle $T $.  We denote the angle between $S_{k-1}$ and $S_k$ as $\theta_{k-1}$. (This angle is well defined since $S_{k-1}$ and $S_k$ intersect in an $m-1$-dimensional affine space.)  As before, we will drop the subscript when it is $k-1$, and just write $\theta$.  Specifically, we define $\theta$ so that  $\cos \theta = \vv^T\w$ (where $\vv$ and $\w$ are defined as in Sec.~\ref{sec:monotonic_chains}, as vectors perpendicular to $S_{k-1} \cap S_k$, and parallel to $S_{k-1}$ and $S_k$, respectively), defining $\theta_k$ similarly for any $k$.  We then express our constraint on the curvature as $\sum_{k=1}^{K-1} | \theta_k | \leq T$.  

Now let $c$ be a constant such that we can bound $\av^T\vv \ge \frac{1}{c}$ for any $k-1$.  To understand this, recall that $\av$ is a unit vector normal to the hyperplane separating $S_{k-1}$ and $S_k$.  By saying this bound holds for all $k-1$, we mean that we are able to choose the hyperplanes that divide the chain into segments so that the angle between the normal to each hyperplane and the following segment is not too big.  We next bound the error in terms of $c$ and $\| \delta \|$.  

Let $\p=\po+\delta$ be as in the last section. We define the embedding error of $\p$ by
\[ E(\p) = \left(B^{(k)} A^{(k)} - X^{(k)T} \right) \p, \]
where $X^{(k)}$ denotes the orthogonal projection to $S_k$, as is used in Sec.~\ref{sec:monotonic_chains}. Noting that, by the construction of our network, $B^{(k)} A^{(k)} \po = X^{(k)T} \po$ (since $\po$ is on $S_k$) and that $X^{(k)T}\delta=0$ (due to the orthonormality of $X^{(k)}$), we obtain
\[ E(\p) = B^{(k)} A^{(k)} \delta. \]
The magnitude of the error therefore is scaled at most by the maximal singular value of $B^{(k)} A^{(k)}$, denoted $\sigma_k$.

To bound $\sigma_k$ we note that $B^{(k)} A^{(k)}=BA+\bv\av^T$ for $k \ge 2$ (where, as before, we drop superscripts so that $B$ denotes $B^{(k-1)})$. Therefore,
\[ \sigma_k \le \sigma_{k-1} + |\av^T\bv|, \]
where $\sigma_{k-1}$ denotes the largest singular value of $BA$. Recall that $\|\av\|=1$ and
\[ \bv = \frac{1}{\av^T\vv}BA(\w-\vv). \]
Note that  $\w-\vv \le \theta_{k-1}$. Therefore,
\[ |\av^T\bv| \le c\sigma_{k-1} \theta_{k-1}, \]
from which we conclude that
\[ \sigma_k \le \sigma_{k-1}(1+c\theta_{k-1}). \]

Finally, note that $B^{(1)} A^{(1)}=X^{(1)T}$, implying that $\sigma_1=1$. We therefore obtain
\[ \sigma_k \le \prod_{j=1}^{k-1} (1 + c\theta_j). \]
Note that $\sum_{j=1}^{k-1} \theta_j \leq T$ and so $\prod_{j=1}^{k-1} (1 + c\theta_j) \le (1 + \frac{cT}{k-1})^{k-1}$. Therefore,
\[ \sigma_k \le \left(1 + \frac{cT}{k-1}\right)^{k-1} \le e^{cT}. \]
We conclude that
\[ \|E(\po+\delta)\| \le e^{cT} \|\delta\|. \]

Many monotonic chains can be divided into segments using hyperplanes in which $c$ is not too big, and may be as low as 1.
For such manifolds, when a point is perturbed away from the manifold, its coordinates will not be changed by more than the magnitude of the perturbation times a small constant factor. For example, if $T=\pi/4$ and $c=1$ then $e_k \le e^{\frac{\pi}{4}} \approx 2.19$. 
Note that  rather than beginning at the start of the monotonic chain, we could "begin" in the middle, and work our way out. That is, provide an orthonormal basis for the middle segment and add hidden units to represent the chain from the central segment toward either ends of the chain. This can reduce the total curvature from the starting point to either end by up to half.  We further emphasize that this bound is not tight. For example, the bound for a single affine segment $(K=1) $ is 1, but since in this case the network encodes an orthogonal projection matrix the actual error is zero.

\section{Combinations of Monotonic Chains}  \label{sec:nonmonotonic-chain}

To handle non-monotonic chains and more general piecewise linear manifolds that can be flattened we show that we can use a network to divide the manifold into monotonic chains, embed each of these separately, and then stitch these embeddings together. 
Suppose we wish to flatten a non-monotonic chain that can be divided into $L$ monotonic chains, $M_1, M_2, ... M_L$. 
Let $A_l$, $\ao_l$ and $B_l$ denote the matrices and bias used to represent the hidden units that flatten $M_l$, which has $K_l$ segments.  We suppose that a set of $J_l$ hyperplanes (that is, a convex polytope) can be found that separate $M_l$ from the other chains.  Let $N_l$ denote a matrix in which the rows represent the normals to these hyperplanes, oriented to point away from $M_l$.  We can concatenate these vertically, letting $A'_l = [A_l; N_l].$   We next let $\Upsilon=-n \mathbf{1}_{m \times J_l}$ where $\mathbf{1}_{m \times J_l}$ denotes an $m \times J_l$ matrix containing all ones and $n$ is a very large constant.  Note that $B_l$ has $m$ rows.  So we can define $B'_l = [B_l ,\Upsilon]$, where the matrices are concatenated horizontally.

We now note that if: 
\[  \uu = B'_l[ A'_l{\x}+\ao_l ]_+  \]
when $\x$ lies on $M_l$, $\uu$ will contain the coordinates of $\x$ embedded in ${\cal R}^m$, as before.  When $\x$ lies on a different monotonic chain, $\uu$ will be a vector with very small negative numbers. Applying RELU will therefore eliminate these numbers.

$A'_l$ and $B'_l$ therefore represent a module consisting of a two layer network that embeds one monotonic chain in ${\cal R}^m$ while producing zero for other chains.  We can then stitch these values together.  First, we must rotate and translate each embedded chain so that each chain picks up where the previous one left off.  Let $R_l$ denote the rotation of each chain, and let $\bo_l$ denote its appropriate translation.  Then, for each chain, the appropriate coordinates are produced by 
\[ [R_l B'_l[ A'_l{\x}+\ao_l ]_+ +\bo_l]_+. \]

We can now concatenate these for all chains to produce the final network.  We let $A$, $\ao$ and $\bo$ be the vertical concatenation of all $A'_l$ and $\ao_l$ and $\bo_l$ respectively, and let $B$ be the block-diagonal concatenation of all $R_lB'_l$. The application of $[B[A\x+\ao]_+ + \bo]_+$ to $\x \in M_l$ will produce a vector with $mL$ entries in which the $m(l-1)+1,...,ml$ entries give the embedded coordinates of $\x$ and the rest of the entries are zero. We can now construct a third layer of the network to then stitch these monotonic chains together.  Let $C$ denote a matrix of size $m \times mL$ obtained by concatenating horizontally $L$ identity matrices of size $m \times m$.  We then describe the output of the network with the equation:
\[  \uu = C[B[ A \x + \ao ]_+ + \bo]_+].  \]
Note, for example, that the first element of $\uu$ is the sum of the first coordinates produced by each module in the first two layers.  Each of these modules produces the appropriate coordinates for points in one monotonic chain, while producing 0 for points in all other monotoinic chains.  

 We note that this summation may result in wrong values if there is overlap between the regions (which will generally be of zero measure). This can be rectified by replacing the summation due to $C$ by max pooling, which allows overlap of any size\footnote{Note that max pooling can easily be implemented with RELU by adding a layer; namely, $\max(x,y)=\frac{1}{2}(x+y-|x-y|)$, where $|x-y|=[x-y]_++[y-x]_+$.}. Together, all three layers will require $\big( \sum_{l=1}^L J_l + m + K_l - 1 \big) +(L+1)m$ units.  If the network is fully connected, this requires $\big( \sum_{l=1}^L J_l + m + K_l - 1 \big)(d+Lm) + Lm^2$ weights.  

Note that the size of this network depends on how many regions are required ($L$) and how many hyperplanes each region needs to separate it from the rest of the manifold ($L_l$).  In the worst case, this can be quite large.  Consider, for example, a 1D manifold that is a polyline that passes through every point with integer coordinates in ${\cal R}^d$.  To separate any portion of this polyline from the rest will require regions that are not unbounded, and so $L_l = O(d)$ for all $l$.  However, such manifolds are somewhat pathological.  We expect that many manifolds can be divided appropriately using many fewer hyperplanes.  We will show this for the example of a Swiss roll and a real world manifold of faces.

\section{Deeper networks and Hierarchical Representations of Manifolds}

We also note that the previously developed constructions can be applied recursively, producing a deeper network that progressively approximates data using linear subspaces of decreasing dimension.  That is, we may first 
divide the data into a set of segments that each lie in a low dimensional subspace whose dimension is higher than the intrinsic dimension of the data.   Then we may subdivide each segment into a set of subsegments of lower dimension, using a similar construction, and deeper layers of the network.  These subsegments may represent the original data, or they be further subdivided by additional layers, until we ultimately produce subsegments that represent the data.  

We first illustrate this hierarchical approach with a simple example that requires only one extra layer in the hierarchy.  Consider a monotonic chain of $K$, $m_2$-dimensional linear segments that collectively lie in a $m_1$-dimensional linear subspace, $\cal L$, of a $d$-dimensional space, with $m_2 < m_1$.  We can construct the first hidden layer with $m_1$ units that are active over the entire monotonic chain, so that their gradient directions form an orthonormal basis for $\cal L$.  The output of this layer will contain the coordinates in ${\cal L}$ of points on the monotonic chain.  These can form the input to two layers that then flatten the chain, as described in Section \ref{sec:monotonic_chains}.  

In Section~\ref{sec:monotonic_chains} we had already shown how to flatten the manifold with two layers that take their input directly from the input space.  Here we accomplish the same end with an extra layer.  However, this construction, while using more layers, may also use fewer parameters.  The construction in Section \ref{sec:monotonic_chains} required $d(m_2 + K - 1)$ parameters.  Our new construction will require $dm_1 + m_1(m_2 + K -1)$ parameters.  Note that as $K$ increases, the number of parameters used in the first construction increases in proportion to $d$, while in the second construction the parameters increase only in proportion to $m_1$.  Consequently, the second construction can be much more economical when $K$ is large and $m_1$ is small.  

In much the same way, we could represent a manifold using a hierarchy of chains.  The first layers can map a $m_1$-dimensional  chain to a linear $m_1$-dimensional output space.  The next layers can select an $m_2$-dimensional chain that lies in this $m_1$-dimensional space, and map it to an $m_2$-dimensional space.  This process can repeat indefinitely, but whether it is economical will depend on the structure of the manifold.

\section{Experiments}

In this section we provide examples of deep networks that illustrate the potential performance of the type of networks that we have described in this paper.  We use two examples.  First, we synthetically generate points on a "Swiss Roll".  We know analytically that this 2D manifold can be flattened to lie in a 2D Euclidean space.  Second, we 
make use of images rendered from a 3D face model under changing viewpoint.  Though high dimensional, these images have only two true degrees of freedom as we alter the elevation and azimuth of the camera.  Consequently, these images can be expected to lie near a 2D manifold.

As the focus of this paper is on the representational capacity of networks, we do not attempt to learn these networks, but rather construct them "by hand."  We make use of prior knowledge of the intrinsic coordinates of each image to divide each set of images into segments.  We then use PCA to fit linear subspaces to each segment, and use the constructions in this paper to build a corresponding neural network that will map these input points to a 2D Euclidean space.


\begin{figure}
\begin{center}
\includegraphics[width=6.7cm]{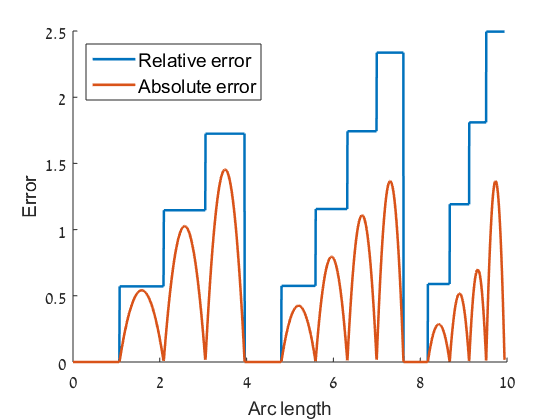}
\end{center}
\caption{These plots show the error in flattening the Swiss Roll. Relative error is constant in every segment, starting from zero  for each monotonic chain and increasing with each segment. The absolute error (for display purposes it is normalized by the maximal distance from the Swiss Roll to its linear approximation) behaves similarly, but vanishes at the end points of each segment where the Swiss Roll and its linear approximation coincide. }
\label{fig:ErrorSwissRoll}
\end{figure}

For the Swiss Roll, as shown in Figure \ref{fig:Overview}, we use hidden units and their corresponding hyperplanes to divide the Roll into three monotonic chains.  We divide each chain into segments, obtaining a total of 14 segments. Figure~\ref{fig:Overview} shows the points that are input into the network, and the 2D representation that the network outputs.  The points are color coded to allow the reader to identify corresponding points. In Figure~\ref{fig:ErrorSwissRoll}  we further plot the absolute and relative error in embedding every point of the Swiss Roll due to the linear approximation used by the network. One can see that the Swiss Roll is unrolled almost perfectly. In fact, despite the relatively large angular extent of each monotonic chain (the three chains range between 126 to 166.5 degrees each in total curvature), the relative error does not exceed 2.5. (In fact, our bound for this case is very loose, amounting to 18.3 for $166.5^\circ$.)  The mean relative error is 0.98.

\begin{figure}
\begin{center}
\includegraphics[width=8.5cm]{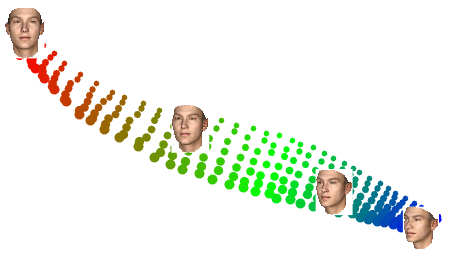}
\end{center}
\caption{The output of a network that approximates images of a face using a monotonic chain.  Each dot represents an image.  They are coded by size to indicate elevation, and color to indicate azimuth.  At four dots, we display the corresponding face images.}
\label{fig:Faces}
\end{figure}

Next we construct a network to flatten a set of images of faces.  We render faces with azimuth ranging from 0 to 35 degrees, and with elevation ranging from 0 to 6 degrees. We use the known viewing parameters to divide these into seven segments, and and then construct a network.  As described at the end of Section \ref{sec:bounds}, we begin with an orthonormal basis for the middle segment of the chain and attach additional segments to both ends of this segment.  The results are shown in Figure \ref{fig:Faces}.  The output does not form a perfect grid, in part because elevation and azimuth need not provide an orthonormal basis for this 2D manifold.  However, we can see that the structure of these variables that describe the input is well-preserved in the output.

\section{Discussion}

The direct technical contribution of this work is to show that deep networks can represent data that lies on a low-dimensional manifold with great efficiency.  In particular, when using a monotonic chain to approximate some component of the data, the addition of only a single neural unit can produce a new linear segment to approximate a region of the data.  This suggests that deep networks may be very effective devices for such dimensionality reduction.  It also may suggest new architectures for deep networks that encourage this type of dimensionality reduction.

We also feel that our work makes a larger point about the nature of deep networks.  It has been shown by \cite{Montufar} that a deep network can divide the input space into a large number of regions in which the network computes piecewise linear functions.  Indeed, the number of regions can be exponential in the number of parameters of the network.  While this suggests a source of great power, it also suggests that there are very strong constraints on the set of regions that can be constructed, and the set of functions that can be computed.  Not every pair of neighboring regions can compute arbitrarily different functions.  Our work shows one way that a single unit can change the linear function that a network computes in two neighboring regions.  We demonstrate that one unit can shape this function to follow a manifold that contains the data.  We feel that this suggests interesting new directions for the study of deep networks.



\section{Acknowledgements}

This research is based upon work supported by the Office
of the Director of National Intelligence (ODNI), Intelligence
Advanced Research Projects Activity (IARPA),
via IARPA R\&D Contract No. 2014-14071600012.  The
views and conclusions contained herein are those of the authors
and should not be interpreted as necessarily representing
the official policies or endorsements, either expressed
or implied, of the ODNI, IARPA, or the U.S. Government.
The U.S. Government is authorized to reproduce and distribute
reprints for Governmental purposes notwithstanding
any copyright annotation thereon.

This research is also based upon work supported by the Israel Binational Science Foundation Grant No. 2010331 
and Israel Science Foundation Grants No. 1265/14.

The authors thank Angjoo Kanazawa and Shahar Kovalsky for their helpful comments.

\bibliography{DeepNetworks}
\bibliographystyle{icml2016mine}

\end{document}